\documentclass[sigconf,nonacm]{aamas} 
\usepackage{balance} % for balancing columns on the final page

\title{Grounding Artificial Intelligence in the Origins of Human Behavior}

\author{Eleni Nisioti}
\affiliation{
  \institution{Flowers Team, Inria and Ensta ParisTech}
  \city{Bordeaux, France}}
\email{eleni.nisioti@inria.fr}

\author{Clément Moulin-Frier}
\affiliation{
  \institution{Flowers Team, Inria and Ensta ParisTech}
  \city{Bordeaux, France}}
\email{clement.moulin-frier@inria.fr}

\begin{abstract}
	Recent advances in Artificial Intelligence (AI) have revived the quest for agents able to acquire an open-ended repertoire of skills. However, although this ability is fundamentally related to the characteristics of human intelligence, research in this field rarely considers the processes that may have guided the emergence of complex cognitive capacities during the evolution of the species. 
	
	Research in Human Behavioral Ecology (HBE) seeks to understand how the behaviors characterizing human nature can be conceived as adaptive responses to major changes in the structure of our ecological niche. In this paper, we propose a framework highlighting the role of environmental complexity in open-ended skill acquisition, grounded in major hypotheses from HBE and recent contributions in Reinforcement learning (RL). We use this framework to highlight fundamental links between the two disciplines, as well as to identify feedback loops that bootstrap ecological complexity and create promising research directions for AI researchers.
	
\end{abstract}

\usepackage{natbib}
\usepackage{xcolor}
\usepackage{url}
\usepackage{graphicx}
\graphicspath{{./images/}}
\usepackage{tikz}
\usetikzlibrary{patterns}
\usetikzlibrary{shapes}
\usetikzlibrary{arrows}
\usetikzlibrary{positioning}
\usepackage{varwidth}

\usepackage{amsmath}               
{   }
\usepackage{amsmath}

\usepackage{amsmath}

\usepackage{amsthm} 
\theoremstyle{definition}

\usepackage[acronym]{glossaries}
\newacronym{hbe}{HBE}{Human Behavioral Ecology}
\newacronym{marl}{MARL}{multi-agent reinforcement learning}

\usepackage[inline]{enumitem}
\usepackage[ruled,linesnumbered]{algorithm2e}
\usepackage{placeins}
\usepackage{makecell}

\date{\today}

\begin{document}
\maketitle

\section{Endless skills most beautiful} 
Be it morphological, behavioral or cultural, the open-endedness of biological life has always been a puzzle for researchers in natural sciences trying to analyze it \cite{darwin,brown_evolutionary_2011} and an inspiration for computational researchers in Artificial Intelligence (AI) trying to implement it \cite{banzhaf_defining_2016,wang_paired_2019}. While a definition of AI has long eluded scientists, it has been proposed that a key property of an intelligent agent may be its ability to adapt to an open-ended set of environments \cite{hernandez,DBLP:journals/corr/abs-1911-01547,lehman_anarchy_2014}. The vigor with which the community is pursuing AI under this new perspective can be measured in the number of related reinforcement learning (RL) sub-fields studying the emergence of behavioral diversity: multi-task, meta-, lifelong and multi-agent reinforcement learning introduce different desiderata but have all studied the design of agents able to efficiently adapt in changing environments.

% Over-fitting and generalization are concepts that lie at the core of machine learning practices. Reinforcement learning (RL) agents have, however, only recently been studied under this light. Deep RL (DRL) has offered generalization over the search space of a given environment, but most works in this area have, until recently, considered a single environment \cite{44806,Vinyals2019GrandmasterLI}. Evolvability, the ability of agents to secure fitness in a current environment without negatively impacting potential benefits in future environments \cite{watson_how_2016}, is an evolutionary concept closely linked to the concept of generalization that is prevalent in supervised learning. The vigor with which the AI community is pursuing this new objective can be measured in the number of closely related disciplines: multi-task, meta and lifelong reinforcement learning introduce different desiderata but are all in essence concerned with the creation of behaviorally diverse RL agents able to act and/or adapt efficiently in unknown environments. 

% \note{Note that I have merged the categorization in the origins proposal with the three pillars of the AI-GA manifesto}

A significant part of past and present RL literature is concerned with the design of new: \begin{enumerate*}[label=(\roman*)]
    \item  algorithms and architectures for learning \cite{al-shedivat_continuous_2018,pmlr-v80-haarnoja18b};
    \item cost functions \cite{berseth_smirl_2020,abdolmaleki_maximum_2018,pmlr-v80-haarnoja18b};
    \item benchmarks and environments \cite{suarez_neural_2019,schwarz_towards_nodate,al-shedivat_continuous_2018,pmlr-v97-cobbe19a}
\end{enumerate*}. The community is gradually approaching a consensus: intelligence is only as general as its environment requires; thus, human-level intelligence should be viewed as constrained by its own ecology rather than as an example of general intelligence \cite{lehman_anarchy_2014,yampolskiy2020human,DBLP:journals/corr/abs-1911-01547}. This has lead to the realization that the problem of defining the environments that intelligent agents are evaluated on is an important bottleneck in advancing research in AI \cite{leibo_autocurricula_2019, clune_ai-gas_2020}.
% This last category signifies a departure from agent-centered research to a more holistic approach and has been termed the problem problem \cite{leibo_autocurricula_2019}. It is slowly becoming clear that the ability of agents to acquire increasingly complex behavior is significantly affected by their ecology \cite{leibo_autocurricula_2019,baker_emergent_2020}. Translating this realization to techniques and best practices for defining the environment will arguably follow the slow and investigative path of scientific research. In this work we pose the question: ``can the AI community leverage existing knowledge to solve the problem problem''?  

\section{Ecology grounds open-endedness}
Despite this progress, research in RL does not often acknowledge that open-ended skill acquisition is fundamentally related to the characteristics of human intelligence \cite{banzhaf_defining_2016}. Couldn't the study of the processes that may have guided the emergence of complex cognitive capacities during the evolution of the human species benefit the study of open-endedness in intelligent agents?

% why do we present hbe?
In this work, we take a step back from our computer-scientific lens and turn our attention towards \acrfull{hbe}, a field concerned with the effect that environmental conditions have had on the evolution of the human species \cite{maslin_synthesis_2015,brown_evolutionary_2011,sear_synthesis_2007}. Works in this field have studied the formation of cooperative groups \cite{chapman_constraints_nodate}, the management of shared resources \cite{gowlett_discovery_2016}, tool use \cite{debeaune_invention_2004}, the emergence of communication \cite{freeberg_social_2012} and culture \cite{origins_tomasello}, whether humans hunt in optimal group sizes \cite{smith_inuit_1985} and how speciation, extinction and dispersal arose in the human history \cite{maslin_synthesis_2015}.

% Researchers in this area  place emphasis on the role of the environment in eliciting the optimal, and therefore most frequently expressed, behavioral phenotypes, without extensive discussion of whether genes, socially learned information or other factors, are responsible for the apparent match between phenotype and environment. 

Admittedly, there are many paths to the acquisition of open-ended skills in AI; grounding our study in human ecology seems to be but one of the options. But there's a number of reasons that may persuade us to explore it:
\begin{enumerate*}[label=(\roman*)]
    \item examining \textit{all} possible ecologies is infeasible considering our modern and foreseeable computational power \cite{clune_ai-gas_2020};
    \item ecologies that are more familiar to ours make it easier to define evaluation criteria. For example, human-ecology inspired metrics such as equality, sustainability and social welfare have been employed to evaluate agents on their ability to forage \cite{perolat_multi-agent_2017}, find optimal taxation strategies \cite{zheng_ai_2020} and play games \cite{baker_emergent_2020};
    \item Darwinian evolution offers an existence proof for human-like open-ended skill acquisition \cite{clune_ai-gas_2020}, as well as empirical data and testable hypotheses;
    \item similar attempts at grounding AI research in a non-computational field have already proven to be a fruitful approach. For example, concepts from Development Science such as intrinsic motivation  \cite{4141061,Pathak_2017_CVPR_Workshops} and embodied language acquisition \cite{cangelosi_integration_2010} have had a significant impact on modern AI research;
    \item the link between \acrshort{hbe} and RL has already been recognized in \cite{frankenhuis_enriching_2019}, where the transfer of ideas has the opposite direction from the one proposed here. A proposal to study major evolutionary transitions in ecology in order to understand the general laws that underlie innovation and transfer insights to artificial evolution is presented in \cite{sole_major_2016}. Our proposal follows a similar direction but focuses on highlighting the overlap between concepts in RL and \acrshort{hbe}. In addition, a number of works in RL have recently resorted to theories from ecology, psychology and economics for inspiration \cite{wang_evolving_2019,hughes_inequity_2018,jaques_social_2019,perolat_multi-agent_2017,koster_model-free_2020}.
\end{enumerate*} 
% One should keep in mind that this approach has as objective creating artificial intelligence that is human-like and by no means general \cite{yampolskiy2020human} 

Our analysis is based on a small, albeit representative, subset of the vast range of hypotheses proposed in \acrshort{hbe} and aims at mapping them to key research questions in RL in order to identify gaps and propose promising research directions. In Section \ref{sec:current}, we provide a brief description of current trends in RL research. Section \ref{sec:hbe} provides a short introduction to the field of \acrshort{hbe}. In Section \ref{sec:main}, we structure our proposed transfer of ecologically-inspired insights based on three recent key research areas in RL: the adaptability of agents, multi-agent dynamics of groups and their cultural repertoire. 
% Be more specific about future direactions: we have found theories that are not being used in ai (the effect of environmental variability, speciation, low-level environmental properties and higher-level propertires are related

\section{Environmental complexity in RL: current trends}\label{sec:current}
The interplay between environmental complexity and open-ended skill acquisition in intelligent agents has been investigated from various perspectives. Below, we point to different works in the areas of single-agent, multi-agent and meta RL.

Single-agent settings have focused on two elements of an agent's learning architecture:\begin{enumerate*}[label=(\roman*)]
\item the neural networks employed for function approximation, whose generalization abilities are an ongoing debate \cite{marcus_rethinking_1998,NIPS2017_10ce03a1}. Observations in \cite{hill_environmental_2020} suggest that the ability of a single neural network to generalize emerges in a complex environment, characterized by multi-modal signals situated in temporally and physically rich spaces that allow for diversity in the agent's perspective;
\item the cost function or type of intrinsic motivation considered. In \cite{berseth_smirl_2020}, useful skills emerge as an agent minimizes future surprise, attempting to counteract the uncertainty inherent in its environment. In curiosity-driven exploration, learning progress generates intrinsic rewards that push an agent to explore and create its own learning curricula\cite{oudeyer_how_2016,colas2020language}.
\end{enumerate*}

% multi-agent RL
In the \acrfull{marl} literature, the automatic discovery of new environments is achieved by multi-agent autocurricula, where environmental complexity arises due to the co-existence of multiple agents \cite{leibo_autocurricula_2019,portelas_automatic_2020,baker_emergent_2020,leibo_malthusian_2018}. In addition to self-play originally used in two-player problem settings \cite{44806}, the presence of multiple agents can give rise to an arms race \cite{baker_emergent_2020} or create population dynamics that lead to the emergence of cooperation \cite{perolat_multi-agent_2017} and exploration \cite{leibo_malthusian_2018}.

% meta-RL
Meta RL aims at equipping agents with the ability to generalize to tasks or environments that have not been encountered during training. Two nested processes of adaptation are traditionally considered: the inner level is a standard RL algorithm operating on a given environment, analog to a developmental learning process. The outer level is tuning the parameters of the inner loop such that it performs well on a distribution of environments, analog to an evolutionary process. Mechanisms are either gradient-based \cite{finn_model-agnostic_2017} or memory-based \cite{wang_learning_2017}.

\begin{figure*}
    \centering
    \scalebox{0.6}{
    \begin{tikzpicture}
\newcommand\xdis{6cm}
\newcommand\ydis{2cm}

 \node[rectangle split, draw, rectangle split parts=2] (env_comp) at (0,0) {\textbf{ \makecell{ Environmental \\ complexity}} \nodepart{two} 
    {\begin{varwidth}{\linewidth}\begin{itemize}
        \item climate variability
        \item resource availability
        \item predator pressure
    \end{itemize}\end{varwidth}}};

\node[rectangle split, draw, rectangle split parts=2] (adapt) at ([xshift=\xdis,yshift=\ydis]env_comp) {\textbf{Adaptability} \nodepart{two} 
    {\begin{varwidth}{\linewidth}\begin{itemize}
        \item learning
        \item speciation
        \item extinction
    \end{itemize}\end{varwidth}}};
  
\node[rectangle split, draw, rectangle split parts=2] (multi) at ([xshift=\xdis,yshift=-\ydis]env_comp) {\textbf{\makecell{Multi-agent \\ dynamics}} \nodepart{two} 
    {\begin{varwidth}{\linewidth}\begin{itemize}
        \item cooperation
        \item competition
        \item stability
    \end{itemize}\end{varwidth}}};

\node[rectangle split, draw, rectangle split parts=2] (techno) at ([xshift=2*\xdis,yshift=1.2*\ydis]env_comp) {\textbf{\makecell{Technology}} \nodepart{two}     {\begin{varwidth}{\linewidth}\begin{itemize}
        \item tool use
        \item engineering
    \end{itemize}\end{varwidth}}};

\node[rectangle split, draw, rectangle split parts=2] (comm) at ([xshift=2*\xdis]env_comp) {\textbf{\makecell{Communication}} \nodepart{two} 
    {\begin{varwidth}{\linewidth}\begin{itemize}
        \item language
        \item grooming
    \end{itemize}\end{varwidth}}};

\node[rectangle split, draw, rectangle split parts=2] (culture) at ([xshift=2*\xdis,yshift=-1.2*\ydis]env_comp) {\textbf{\makecell{Culture}} \nodepart{two}     {\begin{varwidth}{\linewidth}\begin{itemize}
        \item institutions
        \item social norms
        \item religious beliefs
    \end{itemize}\end{varwidth}}};
    
\node[draw, dashed, text centered, minimum height=8.5cm,text width = 0.25\textwidth, text depth = 8 cm, draw,below of=3,anchor=north] (repert) at ([xshift=2*\xdis,yshift=2.5*\ydis]env_comp) {\textbf{Cultural repertoire} };
 %\draw [draw=black!50] (decomp.north west) rectangle +(2.8cm, -2.7cm);

%%% feedforward links %%%
\draw[->, ultra thick] (env_comp)  -- node [near start, yshift=0.5cm] {\cite{frankenhuis_enriching_2019}} (adapt);

\draw[->, ultra thick] (env_comp)  -- node [near start, yshift=0.5cm] {\cite{tomasello_two_2020}} (multi);

\draw[->, ultra thick] ([xshift=-0.5cm]env_comp.north) --  node [near start, xshift=-0.5cm] {\cite{fogarty_niche_2017}}  ([yshift=1.8*\ydis,xshift=-0.5cm]env_comp.north) -| ([xshift=0.5cm]repert.north);

% \draw[->, ultra thick] (multi)  --  node [near start, yshift=0.5cm] {\cite{fogarty_niche_2017}} (techno);
% \draw[->, ultra thick] (multi)  -- node [near start, yshift=0.5cm] {\cite{tomasello_two_2020,dunbar_coevolution_1993}} (comm);
% \draw[->, ultra thick] (multi)  -- node [near start, yshift=0.5cm] {\cite{tomasello_two_2020,dunbar_coevolution_1993}} (culture);

%%% feedback links %%%
\draw[->, ultra thick] ([xshift=-0.5cm]multi.south) -- node [near start,xshift=-0.5cm,yshift=-0.6cm] {\cite{post_eco-evolutionary_2009}} ([xshift=-0.5cm,yshift=-0.3cm]multi.south) -| (env_comp.south); 

\draw[->, ultra thick] (repert.south) -- node [near start,xshift=-0.7cm,yshift=-0.1cm] {\cite{dunbar_coevolution_1993,tomasello_two_2020}} ([yshift=-0.5cm]repert.south) -| (multi.south);

\draw[->, ultra thick] ([xshift=0.5cm]repert.south) -- node [near start,xshift=0.5cm,yshift=-0.2cm] {\cite{fogarty_niche_2017}} ([yshift=-0.7cm,xshift=0.5cm]repert.south) -| ([xshift=-0.5cm]env_comp.south); 

% \draw[->, ultra thick] (comm.west) --  node [near start,xshift=0.5cm] {\cite{cangelosi_integration_2010}} (adapt.east);

\draw[->, ultra thick] ([xshift=-0.5cm]adapt.north) -- node [near start,xshift=-0.5cm,yshift=0.6cm] {\cite{post_eco-evolutionary_2009}} ([xshift=-0.5cm,yshift=0.3cm]adapt.north) -| (env_comp.north); 

\draw[->, ultra thick] (repert.north) -- node [near start,yshift=0.1cm,xshift=-0.5cm] {\cite{cangelosi_integration_2010}} ([yshift=0.5cm]repert.north) -| (adapt.north);

%%% same-level %%%
\draw[<->, ultra thick](adapt.south) -- node [midway,xshift=-0.5cm] {\cite{dunbar_coevolution_1993}} (multi.north);
\draw[<->, ultra thick] (techno.south) -- node [midway,xshift=-0.7cm] {\cite{dunbar_coevolution_1993,tomasello_two_2020}} (comm.north);
\draw[<->, ultra thick] (comm.south) -- node [midway,xshift=-0.7cm] {\cite{dunbar_coevolution_1993,tomasello_two_2020}} (culture.north);

\end{tikzpicture}}
    \caption{Environmental complexity as a main driver in human behavioral ecology. Feed-forward and feedback arrows indicate relationships between the different ecological components, analyzed in the corresponding references from \acrshort{hbe} literature provided as labels.}
    \label{fig:ecology_graph}
\end{figure*}
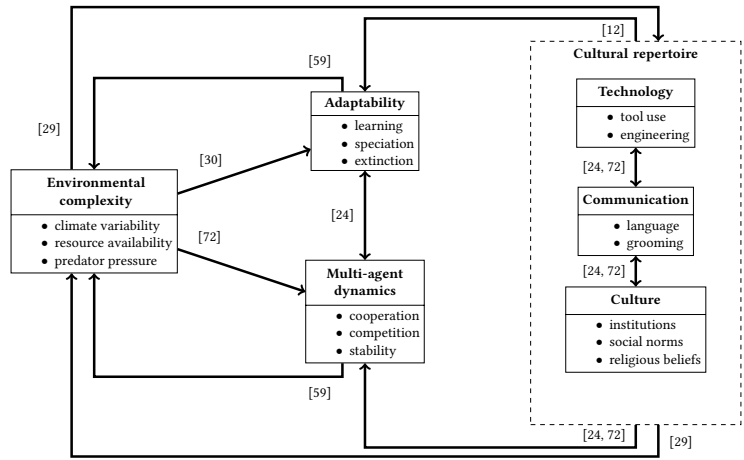

\section{A bird's eye view of Human Behavioral Ecology}\label{sec:hbe}
% explain here the phenotypic gambit and mechanisms of transmission (genes, culture, co-evolution)

\acrshort{hbe} emerged from the field of anthropology in the 70s and is today closely related to evolutionary psychology and cultural evolution \cite{maslin_synthesis_2015}. A key assumption of \acrshort{hbe} is that human behavior is highly flexible and can potentially adapt to changes in its environment \cite{maslin_synthesis_2015,brown_evolutionary_2011}. A profound challenge in studying behavioral adaptation is that the relationship between genes and behavior is to date not clear \cite{rittschof_genomics_2014}. For this reason, \acrshort{hbe} adheres to the assumption of the \textit{phenotypic gambit}, which posits that genetic, psychological or social constraints can be ignored when studying how optimal behaviors arise in a given environment \cite{gambit}.

Which factors contributed to the manifested ability of humans to generalize? What differentiated the human species from others that went extinct due to their inability to adapt to novel environments? These are two important questions that have occupied \acrshort{hbe}, leading to a variety of hypotheses. The spotlight is on the Rift Valley at East Africa approximately 7 million years ago, as it is hypothesized that it constitutes a turning point in our evolutionary trajectory characterized by the first appearance of modern humans and their expansion to other geographical areas \cite{maslin_synthesis_2015}. This evolutionary leap was originally studied under theories that layed emphasis on specific environmental changes. The Savannah hypothesis, for example, suggests that the change in fauna favored bipedal walking, which enabled migration and the creation of new niches for humans \cite{maslin_synthesis_2015}. Later hypotheses under the pulsed climate variability framework, however, suggest that the key change was instead the general environmental complexity characterizing that period \cite{potts_hominin_2013, maslin_synthesis_2015,collard_causes_2020}.

% maybe refer to figure 1
In Figure \ref{fig:ecology_graph}, we introduce a conceptual framework that recognizes important ecological components, as well as the feedforward and feedback links that relate them. In the remainder of this section, we discuss hypotheses studying these relationships in the human ecosystem and, in Section \ref{sec:main}, associate them with research questions in the study of artificial ecosystems. Under the proposed framework, environmental complexity is essentially driven by climate variability, which implies instability in the ecological conditions, in particular through changes in resource availability and exposition to predators \cite{post_eco-evolutionary_2009,frankenhuis_enriching_2019,brown_evolutionary_2011}. This complexity has a strong influence on two major phenomena. First, it drives adaptability both at the evolutionary time scale, through speciation and extinction, \cite{sear_synthesis_2007,potts_hominin_2013} and at the developmental time scale through cognitive mechanisms for learning and abstraction \cite{hougen_evolution_2019,frankenhuis_enriching_2019}. Second, varying the levels of resource availability and exposition to predators has a strong influence on multi-agent dynamics through the modulation of cooperation and competition pressures \cite{dunbar_coevolution_1993,tomasello_understanding_2005}.

% For example, it has been proposed \cite{chapman_constraints_nodate} that the optimal size of primate groups emerges from two opposing constraints: being part of a large group reduces the risk of each individual to be captured by predators, but also increases competition in foraging, reducing the amount of food resources available to each individual. 

The influence of environmental complexity on adaptation and multi-agent dynamics can then have feedback and feedforward effects on the ecological system. First, increased morphological and cognitive complexity due to adaptation, as well as increased complexity in the multi-agent dynamics, feed back to environmental complexity through the modification of resource availability and predation pressure \cite{fogarty_niche_2017,tomasello_two_2020}. For example, the Red Queen hypothesis \cite{doi:https://doi.org/10.1038/npg.els.0001667} proposes that competition among different species is a major drive of evolution, possibly driving an arms race between co-adapting species. Second, adaptation and multi-agent dynamics can bootstrap in a feedforward manner the emergence of more advanced behaviors related to technology (e.g. tool use \cite{debeaune_invention_2004}), communication (e.g. language \cite{tomasello_understanding_2005,freeberg_social_2012}) and culture (e.g. social norms \cite{tomasello_understanding_2005}, institutions \cite{tomasello_understanding_2005} and religions \cite{botero_ecology_2014}). Here again, the emergence of these new behaviors feeds back into environmental complexity through the process of social niche construction \cite{fogarty_niche_2017}, thus creating a positive feedback loop potentially driving the ever-expanding social complexity of human ecology \cite{post_eco-evolutionary_2009,https://doi.org/10.1046/j.1420-9101.2001.00262.x}.

% describe figure and objective
% Having initiated readers to recent trends in RL and \acrshort{hbe}, we next delve deeper into a discussion of the research questions studied by these two fields. 

\section{An ecological perspective on RL}\label{sec:main}
% We will focus our study on three emergent phenomena that have attracted the interest of the RL community: adaptability, cooperative group formation and language. 
% We will make parallels between state-of-the-art questions and insights drawn from ecological literature. 
% In Figure \ref{fig:ecology_graph}, we position the three studies in the interest of different RL communities: meta-RL, multi-agent RL and devrob. We observe that there is an overlap between the interests of these communities and explain these overlaps through an ecological and developmental perspective.
% This taxonomy allows us to extrapolate to future reseach directions.
There seems to be a significant overlap between the questions that RL research poses in its study of the acquisition of open-ended behavior discussed in Section \ref{sec:current} and the hypotheses examined by \acrshort{hbe}, presented in Section \ref{sec:hbe}. In this section, we focus on three emergent phenomena attracting the interest of the RL community: adaptability of individual agents, multi-agent dynamics of groups and their cultural repertoire. By referring to concepts that we originally introduced in Figure \ref{fig:ecology_graph} and highlight in this section, we initiate a dialogue between the two fields and identify key links that we believe deem further investigation by the RL community.

% We review insights from the \acrshort{hbe} literature and attempt to make parallels to research questions asked in RL in order to propose promising directions for future research.

% \begin{figure*}
%     \centering
%     \scalebox{0.7}{
%     \input{tikz/interaction_graph}}
%     %\includegraphics[width=\textwidth]{graph2.png}
%     \caption{}
%     \label{fig:ecology_graph}
% \end{figure*}

\subsection{Adaptability}

%  In the past, RL agents were mainly evaluated based on their ability to adapt to a single task \cite{DBLP:journals/corr/abs-1912-06680,44806,Vinyals2019GrandmasterLI}. Meta reinforcement learning has however shifted the interest towards agents able to learn in multiple environments by developing mechanisms that learn the meta-skill of adaptability.  Mechanisms are either gradient-based \cite{finn_model-agnostic_2017} or memory-based, where the learning procedure is acquired through a recurrent neural network \cite{wang_learning_2017}. An important concept that arose with the need of choosing among potential environments is that of automatic curriculum learning \cite{portelas_automatic_2020}. Among other criteria, environments are chosen based on learning progress, diversity and surprise \cite{portelas_automatic_2020}.

\paragraph{Insights from ecology}  
% Which factors contributed to the manifested ability of humans to efficiently generalize skills and knowledge to novel situations? What differentiated the human species from others that went extinct due to their inability to adapt to novel environments? These are two important questions that have occupied \acrshort{hbe} and have led to a variety of hypothesis. The spotlight is on the Rift Valley at East Africa approximately 7 million years ago, as it is hypothesized that it constitutes a turning point in our evolutionary trajectory characterized by the first appearance of hominins and their expansion to other geographical areas \cite{maslin_synthesis_2015,anton_evolution_2014}. This evolutionary leap was originally studied under theories that layed emphasis on specific environmental changes. The Savannah hypothesis, for example, suggests that the change in fauna favored bipedal walking, which enabled migration and the creation of new niches for humans \cite{maslin_synthesis_2015}. Later hypotheses under the pulsed climate variability framework, however, suggest that the key change was not due to a specific environmental change but due to the general environmental variability characterizing that period \cite{potts_hominin_2013, maslin_synthesis_2015}. In particular, it was the combination of environmental novelty and the transition between periods of ecological stability and instability that lead to adaptability as an evolutionary response.
Under the pulsed climate variability framework discussed in Section \ref{sec:hbe}, environmental factors such as \textbf{climate variability}, \textbf{resource availability} and \textbf{predation pressures} have served as a drive for the ability of humans to adapt to complex environments. Adaptability is achieved through mechanisms whose form depends on properties of the environment. If the environment is constant across time and space, natural selection may favor innate behaviors. By contrast, if the environment varies, natural selection might favor behavioral plasticity: based on environmental observations an agent may be able to switch between different behaviors following innate, and not learned instructions \cite{hougen_evolution_2019,6400847,frankenhuis_enriching_2019}. In cases where the environment changes noticeably across generations but slowly enough within a generation , behavioral plasticity is guided by a process of developmental selection, an example of which is the \textit{learning} process, where an agent's past behavior guide its future behavior. Thus, adaptation to environmental conditions operates on two scales: the evolutionary one drives \textbf{speciation} and \textbf{extinction}, while the developmental one drives \textbf{learning}. Adaptation feeds back into environmental complexity by affecting how environmental changes affect species, equipped with different skill repertoires. For example, during dry periods, the extinction rates of generalist species would reduce as they would be better able to find resources, while specialist species would struggle having lost their environmental niche and their competitive advantage \cite{brown_evolutionary_2011}.

\paragraph{State of the art in RL} 
The outer and inner loop optimization procedure that meta RL algorithms are following matches well with the aforementioned biological mechanisms of adaptation. There is a lot of flexibility in the choice of algorithms used to optimize the two loops: in \cite{10.1145/2001858.2001957,10.1145/3319619.3322044}, an evolutionary algorithm is used in the outer loop and gradient descent in the inner loop, while in \cite{finn_model-agnostic_2017,kirsch_improving_2020} gradient descent is used in both loops. This comes in agreement with recent proposals to view evolution as equivalent to learning \cite{watson_how_2016} and development \cite{fields_does_2020}. Based on ecological insights, we can indicate the following research directions for investigating the effect of environmental variability in adaptability: \begin{enumerate*}[label=(\roman*)]
\item the rate of change is an important hyper-parameter: it should be high enough across generations to trigger mechanisms for adaptability but exhibit similarity patterns in order for innate behaviors to be useful;
\item studying speciation and extinction in groups of intelligent agents may lead to interesting insights in meta RL, which is currently focusing on the continuous adaptation of a single learning mechanism
\end{enumerate*}. 

% how environmental ability affected the cultural repertoire in food-gathering socieites:  Also, the size of the cultural repertoire was
% significantly correlated with some aspect of environmental
% instability, such as risk of resource failure [5,16], mobility
% [16], length of the growing season [3], environmental risk
% [6,14] and effective temperature [15].

\subsection{Multi-agent dynamics}

\paragraph{Insights from ecology}
% 1. Feedforward link: Talk only about small-scale (mutualistic collaboration)
% 2. Describe the dynamics based on Tomasello two key steps when a group is formed it createes the need to maintain cohesion (intra-group dynamics) and defend itself against other groups (inter-group dynamics)
% 3. For the feedback loop search for theories in Eco-evolutionary feedbacks in community and ecosystem ecology. Small-scale collaboration does not probably lead to significant niche construction, it's only when culture and technology (large-scale groups) comes into play that we significantly alter ecology
The emergence of \textbf{cooperation} has given birth to a variety of hypotheses in \acrshort{hbe}. Under the \textit{big mistake hypothesis} \cite{big_mistake}, altruism emerged in small-scale groups due to kin selection or reciprocity, and is today needlessly manifested because evolution has not yet adapted this mechanism. In contrast, the interdependence hypothesis \cite{tomasello_two_2020} proposes a theory for the emergence of cooperation that replaces altruism with mutualistic collaboration. According to it, the need for foraging led to the selective helping of those who were needed as collaborative partners in the future. In sufficiently small groups, social selection was performed based on reputation. The size and structure of groups was dynamically shaped by their need to maintain \textbf{stability} and defend themselves against other groups. \textbf{Competition} between co-existing groups and species also gives rise to arms races, where reciprocal selection and adaptation lead to co-evolution \cite{Benkman2003ReciprocalSC}. Even at this small scale,  the multi-agent dynamics feed back into environmental complexity through process of niche construction: predation, nutrient excretion and habitat modification populations alter their environment and further influence future populations \cite{post_eco-evolutionary_2009}. 
% Under the cultural group selection hypothesis \cite{335276}, which only applies to later stages of human ecology, social groups with more altruists out-competed others. 

% It remains uncertain whether the transmission of the ability to cooperate across generations takes place culturally or through gene-culture co-evolution.

\paragraph{State of the art in RL}
The emergence of cooperation has attracted significant interest in the MARL community. In particular, multi-agent autocurricula have leveraged the feedback effect that multi-agent dynamics have on resource availability \cite{leibo_autocurricula_2019}, as well as arms races between competing groups \cite{baker_emergent_2020}. In \cite{perolat_multi-agent_2017}, agents learn how to fairly access a common pool of resources following simple trial-and-error learning. Recent works have studied the role of intrinsic motivation based on the theories of assortative matching and group selection \cite{wang_evolving_2019}, inequity aversion based on fairness norms \cite{hughes_inequity_2018} and social influence \cite{jaques_social_2019}. In \cite{DBLP:journals/corr/abs-1901-08492}, ecology-inspired hierarchical organizations are used to facilitate decentralized learning. The feedback effect that population dynamics have on the environment was investigated in \cite{leibo_malthusian_2018}, where increase in population size indirectly lead to exploration. As our brief discussion of related \acrshort{hbe} literature however reveals, there exist a number of hypotheses and observations that researchers can leverage to further advance research in MARL:
\begin{enumerate*}
\item according to the inter-dependence hypothesis, the human drive to cooperate was born neither in scenarios that required altruism nor in social dilemmas, which have served as an application ground for the majority of works in MARL. Rather, cooperation arose in Stag hunt type situations, which favored mutualistic collaboration \cite{tomasello_two_2020};
\item group properties such as size and structure are directly related to the multi-agent dynamics of stability and competition. Thus, their influence on the emerging multi-agent autocurricula requires investigation 
\end{enumerate*}.

% Considering the already manifested trend in the RL community of finding inspiration in behavioral sciences, we believe that a deeper understanding of the greater picture of the current ecological framework and a detailed understanding of the different feedforward and feedback links could offer observations that could serve as a research catalyst.

% In particular, the emergence of cooperation should be studied in appropriate game-theoretic models. Intertemporal social dilemmas have dominated the \acrshort{marl} literature \cite{jaques_social_2019,wang_evolving_2019,hughes_inequity_2018}, arguably due to introducing a state-of-the-art challenge. According to \cite{tomasello_two_2020} however, cooperation in small-scale societies arose in Stag Hunt type situations, where all participants have alternatives but would benefit more from coordinating. Thus, \acrshort{marl} researchers can focus on problems that, in hindsight, brought the biggest evolutionary leaps.

% put this in last section \item the group size should be explicitly taken into account. To the best of our knowledge, no works in \acrshort{marl} have investigated the interplay between group size and cooperation, although ecology suggests that different mechanisms are required at different scales and that cooperation followed an evolutionary trajectory (instead of being independently reinvented at different stages of the human evolution). We believe that the study of complex skills, such as foraging, tool use and language should take the group size dynamics into account.

\subsection{Cultural repertoire}

\paragraph{Insights from ecology}  

Non-human species often exhibit impressive behavioral repertoires \cite{krams_linking_2012}. However, human ecology is characterized by a uniquely large behavioral repertoire: \textbf{engineering}, \textbf{language}, \textbf{social norms}, \textbf{institutions} and \textbf{religious beliefs} constitute a complex cultural ecosystem that has lead scientists on the search of factors that differentiated us from other species \cite{tomasello_two_2020,tomasello_becoming_2019,botero_ecology_2014}. According to the inter-dependence hypothesis, social norms and institutions emerged to counteract the fact that reputation alone could no longer alleviate the problem of free riding in large groups. In addition, the social complexity hypothesis \cite{krams_linking_2012} states that language worked as a bonding mechanism that replaced grooming, practiced in small-scale societies, and thus helped with maintaining group stability in larger groups \cite{dunbar_coevolution_1993}. The feedforward and feedback links associated with \textbf{tool use} have also been investigated under a number of, often contesting, hypotheses. Based on the data analysis in \cite{fogarty_niche_2017}, environmental variability such as risk of resource failure, mobility and climate characteristics correlate significantly with tool use in food-gathering societies. However, it is the group size and not these factors that affect tool use in food-producing societies. It is therefore conjectured that the feedback link of societies with a larger cultural repertoire has a stabilizing effect, dampening the forward impact of environmental variability \cite{collard_niche_nodate}.

% The social complexity hypothesis also associates group size and cognitive capacity: while the minimum group size is determined by ecological conditions, such as the presence of resources and predators, the maximum group size beyond which stability is lost is determined by the social structure, with constructs such as language and institutions serving as a catalyst for improving the connectivity of individuals \cite{dunbar_coevolution_1993}. 

% The study of language understanding in AI has until now followed a primarily data-driven approach, with recent deep learning models exhibiting impressive performance by leveraging unstructured, unlabeled data \cite{bisk_experience_2020}. Recent concerns however about the viability of this approach, have shifted the interest towards associating language with real-world perception \cite{bisk_experience_2020}. This realization has been extensively studied in the DevRob community, where embodied cognitive systems develop communication skills by integrating the development of action and social interaction \cite{cangelosi_integration_2010}. 

Another important link at this level is the relationship between tool use, language and adaptability. Studies of biological motor systems and language acquisition in infants have revealed that action and language representation share a similar compositional structure \cite{cangelosi_integration_2010}. To understand how this similarity between two apparently distinct systems arose, one needs to turn to the origins of this relationship in human ecology. According to the Corballis hypothesis \cite{corballis}, the ability of primates' to manipulate tools may have played a pivotal role in the evolution of language by creating the cognitive representations that compositionality requires. At the same time, the compositional structure of language is hypothesized to be an enabler of flexible and adaptable behavior, thus feeding back to adaptability \cite{cangelosi_integration_2010}.

\paragraph{State of the art in RL}
Recent works in \acrshort{marl} have studied the emergence of communication \cite{lazaridou_emergent_2020,foerster_learning_2016} as well as social norms and conventions \cite{koster_model-free_2020,10.5555/2936924.2937056}. The effect that the type of social network organization, its average degree, and local connectivity has on communication learning in groups of deep reinforcement learning agents is investigated in  \cite{dubova_reinforcement_nodate}. This work constitutes an important first step in the realm of the social complexity hypothesis \cite{dunbar_coevolution_1993}, but there remain a number of research directions lying at the intersection of \acrshort{marl} and meta RL:
\begin{enumerate*}
\item the feedback effect that the cultural repertoire has on environmental variability through cultural niche construction \cite{fogarty_niche_2017} can potentially create more powerful autocurricula than the already studied ones based on niche construction in small-scale groups \cite{baker_emergent_2020,perolat_multi-agent_2017};
\item studying the stabilization effect of cultural niche construction can provide important insights to the problem of scaling up artificial multi-agent systems ;
\item the relationship between action/language compositionality and the ability of agents to generalize and adapt needs to be further investigated in order to transfer insights from human language acquisition to intelligent agents \cite{colas2020language}
\end{enumerate*}.

\section{Discussion}

Our proposal is but a preliminary step towards realizing the potential of a cross-disciplinary dialogue between the \acrshort{hbe} and RL communities, a glimpse of which has already been offered by recent works. For example, it has been proposed that RL methods can enable experimentation in large state spaces with unknown dynamics, overcoming the limitations of stochastic dynamic programming approaches currently dominating \acrshort{hbe} \cite{frankenhuis_enriching_2019}. The family of RL methods is, however, constantly expanding, with \acrshort{marl} in particular offering insights into emergent complexity. Thus, our discussion reveals that the potential of RL as a computational tool for enriching the analytical toolbox of \acrshort{hbe} has not been fully realized. This potential has already been recognized for language development, an important component of human ecology \cite{moulin-frier_multi-agent_2020}. 

At the other end of the spectrum, recent works in RL are increasingly drawing inspiration from ecology \cite{wang_evolving_2019}, as well as psychology \cite{jaques_social_2019} and economics \cite{hughes_inequity_2018}. As our proposal illustrates, however, this attempt is currently limited due to the lack of an overall picture and a thorough examination of feedforward and feeback links taking place at different ecological levels. We believe that conceptual frameworks, such as the one that guided our current analysis in Figure \ref{fig:ecology_graph}, can serve as an important basis for this inter-disciplinary dialogue, with different questions zooming in on different sub-parts and potentially revealing lower-level relationships. 

% At the same time, RL and evolution in CS are continuously experiencing an evolution that needs to be appreciated from the perspective of natural sciences \cite{hougen_evolution_2019}.

% The proposed dialogue aims beyond a unilateral enrichment of the two fields. As has been recognized for properties such as open-endedness \cite{banzhaf_defining_2016}, the acquisition of open-ended behavior is essentially a generic property of both natural and artificial agents, making the conception of an integrating framework an important step towards understanding and implementing it.  

\section*{Acknowledgement}
This work is funded by the Inria Exploratory Action ORIGINS (2020-2022) and the ANR JCJC ECOCURL project (2020-2025).

\bibliographystyle{ACM-Reference-Format} 
\bibliography{bibliography}

\end{document}